# Policy Improvement for POMDPs using Normalized Importance Sampling

Christian R. Shelton
Artificial Intelligence Lab
Massachusetts Institute of Technology
Cambridge, MA 02139
cshelton@ai.mit.edu

## Abstract

We present a new method for estimating the expected return of a POMDP from experience. The estimator does not assume any knowledge of the POMDP, can estimate the returns for finite state controllers, allows experience to be gathered from arbitrary sequences of policies, and estimates the return for any new policy. We motivate the estimator from function-approximation and importance sampling points-of-view and derive its bias and variance. Although the estimator is biased, it has low variance and the bias is often irrelevant when the estimator is used for pair-wise comparisons. We conclude by extending the estimator to policies with memory and compare its performance in a greedy search algorithm to the REINFORCE algorithm showing an order of magnitude reduction in the number of trials required.

## 1 Introduction

We assume a standard reinforcement learning setup: an agent interacts with an environment modeled as a partially-observable Markov decision process. Consider the situation after a sequence of interactions. The agent has now accumulated data and would like to use that data to select how it will act next. In particular, it has accumulated a sequence of observations, actions, and rewards and it would like to select a policy, a mapping from observations to actions, for future interaction with the world. Ultimately, the goal of the agent is to find a policy mapping that maximizes the agent's return, the sum of rewards experienced.

[Kearns et al., 1999] presents a method for estimating the return for every policy simultaneously using data gathered while executing a fixed policy. In this paper we consider the case where the policies used for gathering data are unrestricted. Either we did not have control over the method for data collection, or we would like to allow the learning algorithm the freedom to pick any policy for any trial and still be able to use the data.

Importance sampling has been studied before in conjunction with reinforcement learning. In particular, [Precup et al., 2000, Precup et al., 2001] use importance sampling to estimate Q-values for MDPs with function approximation for the case where all data have been collected using a single policy. [Meuleau et al., 2001] uses importance sampling for POMDPs, but to modify the REINFORCE algorithm [Williams, 1992] which ignores past trials. [Peshkin and Mukherjee, 2001] considers estimators very similar to the ones developed here and prove theoretical PAC bounds for them. This paper differs from previous work in that it allows multiple sampling policies, uses normalized estimators for POMDP problems, derives exact bias and variance formulas for normalized and unnormalized estimators, and extends importance sampling from reactive policies to finite state controllers.

In the next section we develop two estimators (unnormalized and normalized). Section 3 shows that while the normalized estimator is biased, its variance is much lower than the unnormalized (unbiased) estimator resulting in a better estimator for comparisons. Section 4 demonstrates some results on simulated environments. We conclude with a discussion of how to improve the estimator further.

## 2 Estimators

### 2.1 Notation

In this paper we will use $s$ to represent the hidden state of the world, $x$ for the observation, $a$ for the action, and $r$ for the reward. Subscripts denote the time index and superscripts the trial number. We will be studying episodic tasks of fixed-length, $T$. The starting distribution over states is fixed (and unknown).

Let $\pi(x, a)$ be a policy (the probability of picking action $a$ upon observing $x$). For the moment we will consider only reactive policies of this form. $h$ represents a history[1] (of $T$ time steps) and therefore is a tuple of four sequences: states ($s_1$ through $s_T$), observations ($x_1$ through $x_T$), actions ($a_1$ through $a_T$), and rewards ($r_1$ through $r_T$). The state sequence is not available to the algorithm and is for theoretical consideration only. Lastly, we let $R$ be the return (or sum of $r_1$ through $r_T$).

$\pi^1$ through $\pi^n$ are the $n$ policies tried. $h^1$ through $h^n$ are the associated $n$ histories with $R^1$ through $R^n$ being the returns of those histories. Thus during trial $i$ the agent executed policy $\pi^i$ resulting in the history $h^i$. $R^i$ is used as a shorthand notation for $R(h^i)$, the return of $h^i$.

## 2.2 Importance Sampling

Importance sampling is typically presented as a method for reducing the variance of the estimate of an expectation by carefully choosing a sampling distribution [Rubinstein, 1981]. For example, the most direct method for evaluating $\int f(x)p(x)\,dx$ is to sample i.i.d. $x_i \sim p(x)$ and use $\frac{1}{n}\sum_i f(x_i)$ as the estimate. However, by choosing a different distribution $q(x)$ which has higher density in the places where $|f(x)|$ is larger, we can get a new estimate which is still unbiased and has lower variance. In particular, we now draw $x_i \sim q(x)$ and use $\frac{1}{n}\sum_i f(x_i)\frac{p(x_i)}{q(x_i)}$ as our estimate. This can be viewed as estimating the expectation of $f(x)\frac{p(x)}{q(x)}$ with respect to $q(x)$ which is like approximating $\int f(x)\frac{p(x)}{q(x)}q(x)\,dx$ with samples drawn from $q(x)$. If $q(x)$ is chosen properly, our new estimate has lower variance. It is always unbiased provided that the support of $p(x)$ and $q(x)$ are the same. In this paper we only consider stochastic policies that have a non-zero probability of taking any action at any time. Therefore, our sampling and target distributions will always have the same support.

Instead of choosing $q(x)$ to reduce variance, we will be forced to use $q(x)$ because of how our data was collected. Unlike the traditional setting where an estimator is chosen and then a distribution is derived which will achieve minimal variance, we have a distribution chosen and we are trying to find an estimator with low variance.

## 2.3 Sampling Ratios

We have accumulated a set of histories ($h^1$ through $h^n$) each recorded by executing a (possibly different) policy ($\pi^1$ through $\pi^n$). We would like to use this data to find a guess at the best policy.

A key observation is that we can calculate one factor in the probability of a history given a policy. In particular, that probability has the form

$$p(h|\pi) = p(s_1) \prod_{t=1}^{T} p(x_t|s_t)\pi(x_t,a_t)p(s_{t+1}|s_t,a_t)$$
$$= \left[ p(s_1) \prod_{t=1}^{T} p(x_t|s_t)p(s_{t+1}|s_t,a_t) \right] \left[ \prod_{t=1}^{T} \pi(x_t,a_t) \right]$$
$$= W(h)A(h,\pi) \ .$$

$A(h,\pi)$, the effect of the agent, is computable whereas $W(h)$, the effect of the world, is not because it depends on knowledge of the underlying state sequence. However, $W(h)$ does not depend on $\pi$. This implies that the ratios necessary for importance sampling are exactly the ratios that are computable without knowing the state sequence. In particular, if a history $h$ was drawn according to the distribution induced by $\pi$ and we would like an unbiased estimate of the return of $\pi'$, then we can use $R(h)\frac{p(h|\pi')}{p(h|\pi)}$ and although neither the numerator nor the denominator of the importance sampling ratio can be computed, the $W(h)$ term in each cancels leaving a ratio of $A(h,\pi')$ to $A(h,\pi)$ which can be calculated. A different statement of the same fact has been shown before in [Meuleau et al., 2001]. This fact will be exploited in each of the estimators in this paper.

## 2.4 Importance Sampling as Function Approximation

Because each $\pi^i$ is potentially different, each $h^i$ is drawn according to a different distribution and so while the data are drawn independently, they are not identically distributed. This makes it difficult to apply importance sampling directly. The most obvious thing to do is to construct $n$ estimators (one from each data point) and then average them. This estimator has the problem that its variance can be quite high. In particular, if only one of the sampled policies is close to the target policy, then only one of the elements in the sum will have a low variance. The other variances will be very high and overwhelm the total estimate. We might then only use the estimate from the policy that is most similar to the target policy. Yet, we would hope to do better by using all of the data.

---
[1] It might be better to refer to this as a trajectory since we will not limit $h$ to represent only sequences that have been observed; it can also stand for sequences that might be observed. However, the symbol $t$ is over used already. Therefore, we have chosen to use $h$ to represent state-observation-action-reward sequences.

To motivate the estimator of the next section, we demonstrate how importance sampling can be viewed in terms of function approximation. Importance sampling in general seeks to estimate $\int f(x)p(x)\,dx$. Consider estimating this integral by evaluating $\int \hat{f}(x)\hat{p}(x)\,dx$ where $\hat{f}$ and $\hat{p}$ are approximations of $f$ and $p$ derived from data. In particular, with a bit of foresight we will choose $\hat{f}$ and $\hat{p}$ to be nearest-neighbor estimates. Let $i(x)$ be the index of the data point nearest to $x$. Then,

$$\hat{f}(x) = f(x^{i(x)})$$
$$\hat{p}(x) = p(x^{i(x)}) \ .$$

We now must define the size of the "basin" near sample $x^i$. In particular we let $\alpha^i$ be the size of the region of the sampling space closest to $x^i$. In the case where the sampling space is discrete, this is the number of points which are closer to sampled point $x^i$ than any other sampled point. For continuous sampling spaces, $\alpha^i$ is the volume of space which is closest to $x^i$. With this definition,

$$\int \hat{f}(x)\hat{p}(x)\,dx = \sum_i \alpha^i f(x^i) p(x^i) \ .$$

$\alpha^i$ cannot be computed and thus we will need to approximate it. Let $q(x)$ be the distribution from which $x^i$ was sampled. On average, we expect the density of points to be inversely proportional to the volume nearest each point. For instance, if we have sampled uniformly from a unit volume and the average density of points is $d$, then the average volume nearest any given point is $\frac{1}{d}$. Extending this principle, we will take the estimate of $\alpha^i$ to be inversely proportional to the sampling density at $x^i$. This yields the standard importance sampling estimator

$$\frac{1}{n} \sum_i f(x^i) \frac{p(x^i)}{q(x^i)} \ .$$

More importantly, this derivation gives insight into how to merge samples from different distributions, $q^1(x)$ through $q^n(x)$. Not until the estimation of $\alpha^i$ did we require knowledge about the sampling density. We can use the same approximations for $\hat{f}$ and $\hat{p}$. When estimating $\alpha^i$ we need only an estimate of the density of points at $\alpha^i$ to estimate the volume near $x^i$. We therefore take the mixture density, $\frac{1}{n}\sum_i q^i(x)$ (the average of all of the sampling densities) as the distribution of points in sample space. Applying this change results in the estimator

$$\sum_i f(x^i) \frac{p(x^i)}{\sum_j q^j(x^i)} \ .$$

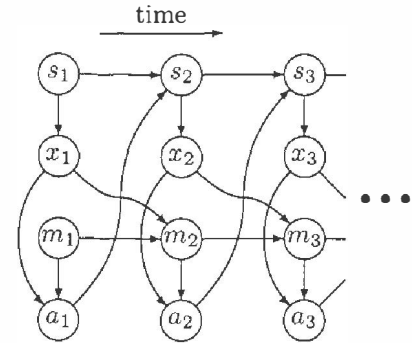

Figure 1: Dependency graph for agent-world interaction with memory model

which, when translated to the POMDP estimation problem, becomes

$$\sum_{i=1}^n R^i \frac{p(h^i|\pi)}{\sum_{j=1}^n p(h^i|\pi^j)} \ . \qquad (1)$$

This estimator is unbiased (shown in the full version of this paper [Shelton, 2001]) and has a lower variance than the sum of $n$ single sample estimators because if one of the sampling distributions is near the target distribution, then all elements in the sum share the benefit.

## 2.5 Normalized Estimates

We can normalize the importance sampling estimate to obtain a lower variance estimate at the cost of adding bias. Previous work has used a variety of names for this including weighted uniform sampling [Rubinstein, 1981], weighted importance sampling [Precup et al., 2000], and ratio estimate [Hesterberg, 1995]. In general, such an estimator has the form

$$\frac{\sum_i f(x^i) \frac{p(x^i)}{q(x^i)}}{\sum_i \frac{p(x^i)}{q(x^i)}} \ .$$

The problem with the previous estimator can be seen by noting that the function approximator $\hat{p}(h)$ does not integrate (or sum) to 1. Instead of $\hat{p} = p(x^{i(x)})$, we make sure $\hat{p}$ integrates (or sums) to 1: $\hat{p} = p(x^{i(x)})/Z$ where $Z = \sum_i \alpha^i p(x^i)$. When recast in terms of our POMDP problem the estimator is

$$\frac{\sum_{i=1}^n R^i \frac{p(h^i|\pi)}{\sum_{j=1}^n p(h^i|\pi^j)}}{\sum_{i=1}^n \frac{p(h^i|\pi)}{\sum_{j=1}^n p(h^i|\pi^j)}} \ . \qquad (2)$$

## 2.6 Adding Memory

So far we have only discussed estimators for reactive policies (policies that map the immediate observation to an action). We would like to be able to also estimate the return for policies with memory. Consider adding memory in the style of a finite-state controller. At each time step, the agent reads the value of the memory along with the observation and makes a choice about which action to take and the new setting for the memory. The policy now expands to the form $\pi(x, m, a, m') = p(a, m'|x, m)$, the probability of picking action $a$ and new memory state $m'$ given observation $x$ and old memory state $m$. Now let us factor this distribution, thereby slightly limiting the class of policies realizable by a given memory size but making the model simpler. In particular we consider an agent model where the agent's policy has two parts: $\pi_a(x, m, a)$ and $\pi_m(x, m, m')$. The former is the probability of choosing action $a$ given that the observation is $x$ and the internal memory is $m$. The latter is the probability of changing the internal memory to $m'$ given the observation is $x$ and the internal memory is $m$. Thus $p(a, m'|x, m) = \pi_a(x, m, a)\pi_m(x, m, m')$. By this factoring of the probability distribution of action-memory choices, we induce the dependency graph shown in figure 1.

If we let $M$ be the sequence $m_1, m_2, \ldots, m_T$, $p(h|\pi)$ can be written as

$$\sum_M p(h, M|\pi)$$

$$= \sum_M p(s_1)p(m_1) \prod_{t=1}^T p(x_t|s_t)\pi_a(x_t, m_t, a_t)$$

$$\pi_m(x_t, m_t, m_{t+1})p(s_{t+1}|s_t, a_t)$$

$$= \left[p(s_1) \prod_{t=1}^T p(x_t|s_t)p(s_{t+1}|s_t, a_t)\right]$$

$$\left[\sum_M p(m_1) \prod_{t=1}^T \pi_a(x_t, m_t, a_t)\pi_m(x_t, m_t, m_{t+1})\right]$$

$$= W(h)A(h, \pi) ,$$

once again splitting the probability into two factors: one for the world dynamics and one for the agent dynamics. The $A(h, \pi)$ term involves a sum over all possible memory sequences. This can easily be computed by noting that $A(h, \pi)$ is the probability of the action sequence given the observation sequence where the memory sequence is unobserved. This is a (slight) variation of a hidden Markov model: an input-output HMM. The difference is that the HMM transition and observation probabilities for a time step (the memory policy and the action policy respectively) depend on the value of $x$ at that time step. Yet, the $x$'s are visible making it possible to compute the probability and its derivative by using the forward-backward algorithm.

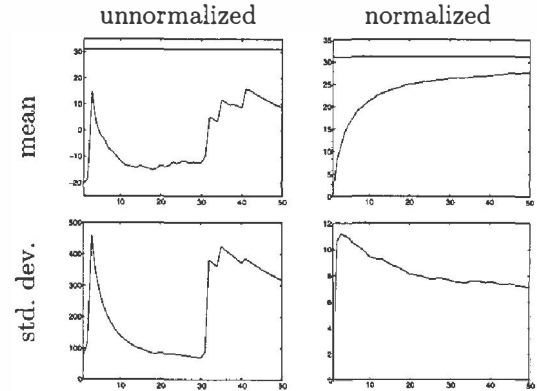

Figure 2: Empirical estimates of the means and standard deviations for the unnormalized and normalized estimates of the return differences as a function of the number of data points. The data were collected by executing the policy corresponding to the point $(0.4, 0.6)$ in figure 3. The estimators were evaluated at the policies $(0.3, 0.9)$ and $(0.4, 0.5)$. Plotted above are the means and standard deviations of these estimates averaged over 600 experiments. The horizontal line on the plots of the mean represent the true difference in returns. The normalized plots fit the theoretical values well. The unnormalized plots demonstrate that even 600 trials are not enough to get a good estimate of the bias or variance. The unnormalized mean should be constant at the true return difference and the standard deviation should decay as $\frac{1}{\sqrt{n}}$. However, because the unnormalized estimator is much more asymmetric (it relies on a few very heavily weighted unlikely events to offset the more common events), the graph does not correspond well to the theoretical values. This is indicative of the general problem with the high variance unnormalized estimates.

As such, we can now use the same estimators and allow for policies with memory. In particular, the estimator has explicit knowledge of the working of the memory. This is in direct contrast to the method of adding the memory to the action and observation spaces and running a standard reinforcement learning algorithm where the agent must learn dynamics of its own memory. With our explicit memory model, the learning algorithm understands that the goal is to produce the correct action sequence and uses the memory state to do so by coordinating the actions in different time steps.

## 3 Estimator Properties

It is well known that importance sampling estimates (both normalized and unnormalized) are consistent [Hesterberg, 1995, Geweke, 1989,

Kloek and van Dijk, 1978]. Additionally, normalized estimators have smaller asymptotic variance if the sampling distribution does not exactly match the distribution to estimate [Hesterberg, 1995]. However, we are more interested in the case of finite sample sizes.

The estimator of equation 1 is unbiased. That is, for a set of chosen policies, $\{\pi^1, \pi^2, \ldots, \pi^n\}$, the expectation over experiences of the estimator evaluated at $\pi$ is the true expected return for executing policy $\pi$. Similarly, the estimator of section 2.5 (equation 2) is biased. In specific, it is biased towards the expected returns of $\{\pi^1, \pi^2, \ldots, \pi^n\}$.

The goal of constructing these estimators is to use them to choose a good policy. This involves comparing the estimates for different values of $\pi$. Therefore instead of considering a single point we will consider the difference of the estimator evaluated at two different points, $\pi_A$ and $\pi_B$. In other words, we will use the estimators to calculate an estimate of the difference in expected returns between two policies. The appendix of the full version of this paper [Shelton, 2001] details the derivation of the biases and variances. We only quote the results here. These results are for using the same data for both the estimates at $\pi_A$ and $\pi_B$.

We denote the difference in returns for the unnormalized estimator as $D_U$ and the difference for the normalized estimator as $D_N$. First, a few useful definitions (allowing $R_X = E[R|\pi_X]$):

$$\bar{p}(h) = \frac{1}{n}\sum_i p(h|\pi^i)$$

$$\tilde{p}(h,g) = \frac{1}{n}\sum_i p(h|\pi^i)p(g|\pi^i)$$

$$b_{A,B} = \iint [R(h) - R(g)]\frac{p(h|\pi_A)p(g|\pi_B)}{\bar{p}(h)\bar{p}(g)}\tilde{p}(h,g)\,dh\,dg$$

$$s^2_{X,Y} = \int R^2(h)\frac{p(h|\pi_X)p(h|\pi_Y)}{\bar{p}(h)}\,dh$$

$$\overline{s^2_{X,Y}} = \int (R(h) - R_X)(R(h) - R_Y)\frac{p(h|\pi_X)p(h|\pi_Y)}{\bar{p}(h)}\,dh$$

$$\eta^2_{X,Y} = \iint R(h)R(g)\frac{p(h|\pi_X)p(g|\pi_Y)}{\bar{p}(h)\bar{p}(g)}\tilde{p}(h,g)\,dh\,dg$$

$$\overline{\eta^2_{X,Y}} = \iint (R(h) - R_X)(R(g) - R_Y)$$
$$\frac{p(h|\pi_X)p(g|\pi_Y)}{\bar{p}(h)\bar{p}(g)}\tilde{p}(h,g)\,dh\,dg$$

Note that all of these quantities are invariant to the number of samples provided that the relative frequencies of the sampling policies remains fixed. $\bar{p}$ and $\tilde{p}$ are measures of the average distribution over histories. $s^2_{X,Y}$ and $\eta^2_{X,Y}$ are measures of second moments and $\overline{s^2_{X,Y}}$ and $\overline{\eta^2_{X,Y}}$ are measures of (approximately) centralized second moments. $b_{A,B}$ is the bias of the normalized estimate of the return difference.

The means and variances are[2]

$$E[D_U] = R_A - R_B$$
$$E[D_N] = R_A - R_B - \frac{1}{n}b_{A,B}$$
$$\mathrm{var}[D_U] = \frac{1}{n}(s^2_{A,A} - 2s^2_{A,B} + s^2_{B,B})$$
$$\quad - \frac{1}{n}(\eta^2_{A,A} - 2\eta^2_{A,B} + \eta^2_{B,B})$$
$$\mathrm{var}[D_N] = \frac{1}{n}(\overline{s^2_{A,A}} - 2\overline{s^2_{A,B}} + \overline{s^2_{B,B}})$$
$$\quad - \frac{1}{n}(\overline{\eta^2_{A,A}} - 2\overline{\eta^2_{A,B}} + \overline{\eta^2_{B,B}})$$
$$\quad - 3\frac{1}{n}(R_A - R_B)b_{A,B} + O(\frac{1}{n^2})\ .$$

The bias of the normalized return difference estimator and the variance of both return differences estimators decrease as $\frac{1}{n}$. It is useful to note that if all of the $\pi^i$'s are the same, then $\tilde{p}(h,g) = \bar{p}(h)\bar{p}(g)$ and thus $b_{A,B} = R_A - R_B$. In this case $E[D_N] = \frac{n-1}{n}(R_A - R_B)$. If the estimator is only used for comparisons, this value is just as good as the true return difference (of course, for small $n$, the same variance would cause greater relative fluctuations).

In general we expect $b_{A,B}$ to be of the same sign as $R_A - R_B$. We would also expect $\overline{s^2_{X,Y}}$ to be less than $s^2_{X,Y}$ (and similarly $\overline{\eta^2_{X,Y}}$ to be less than $\eta^2_{X,Y}$). $\overline{s^2_{X,Y}}$ and $\overline{\eta^2_{X,Y}}$ depend on the difference of the returns from the expected return under $\pi_X$ and $\pi_Y$. $s^2_{X,Y}$ and $\eta^2_{X,Y}$ depend on the difference of the returns from zero. Without any other knowledge of the underlying POMDP, we expect that the return from an arbitrary history be closer to $R_A$ or $R_B$ than the arbitrarily chosen value 0. If $b_{A,B}$ is the same sign as the true difference in returns and the overlined values are less than their counterparts, then the variance of the normalized estimator is less than the variance of the unnormalized estimator. These results are demonstrated empirically in figure 2 where we compared the estimates for the problem described in section 4.2.

---

[2] For the normalized difference estimator, the expectations shown are for the numerator of the difference. The denominator is a positive quantity and can be scaled to be approximately 1. Because the difference is only used for comparisons, this scaling makes no difference in its performance. See the full version [Shelton, 2001] for more details.

## 4 Experiments

### 4.1 Reinforcement Learning Algorithm

We can turn either of these estimators into a greedy learning algorithm. To find a policy by which to act, the agent maximizes the value of the estimator by hill-climbing in the space of policies (using the previous policy as a starting point) until it reaches a maximum. The agent uses this new policy for the next trial. After the trial, it adds the new policy-history-return triple to its data and repeats.

The hill-climbing algorithm must be carefully chosen. For many estimates, the derivative of the estimate varies greatly in magnitude (as shown below). Therefore, we have found it best to use the direction of the gradient, but not its magnitude to determine the direction in which to climb. In particular, we employ a conjugate gradient descent algorithm using a golden-ratio line search [Press et al., 1992].

### 4.2 Two-dimensional Problem

Figure 3 shows a simple world for which the policy can be described by two numbers (the probability of going left when in the left half and the probability of going left when in the right half) and the true expected return as a function of the policy. Figure 4 compares the normalized (equation 1) and unnormalized (equation 2) estimators both with the greedy policy improvement algorithm and under random policy choices. We feel this example is illustrative of the reasons that the normalized estimate works much better on the problems we have tried. Its bias to observed returns works well to smooth out the space. The estimator is willing to extrapolate to unseen regions where the unnormalized estimator is not. This also causes the greedy algorithm to explore new areas of the policy space whereas the unnormalized estimator gets trapped in the visited area under greedy exploration and does not successfully maximize the return function.

### 4.3 Twenty-dimensional Problem

Although the left-right problem was nice because the estimators could be plotted, it is very simple. The load-unload problem of figure 5 is more challenging. To achieve reasonable performance, the actions must depend on the history. We give the agent one memory bit as in section 2.6; this results in twenty independent policy parameters. REINFORCE [Williams, 1992] has also been used to attack a very similar problem [Peshkin et al., 1999]. We compare the results of the normalized estimator with greedy search to REIN-

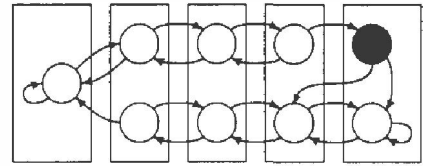

Figure 5: Diagram of the "load-unload" world. This world has nine states. The horizontal axis corresponds to the positioning of a cart. The vertical axis indicates whether the cart is loaded. The agent only observes the position of the cart (five observations denoted by boxes). The cart is loaded when it reaches the left-most state and if it reaches the right-most position while loaded, it is unloaded and the agent receives a single unit of reward. The agent has two actions at each point: move left or move right. Moving left or right off the end leaves the cart unmoved. Each trial begins in the left-most state and lasts 100 time steps.

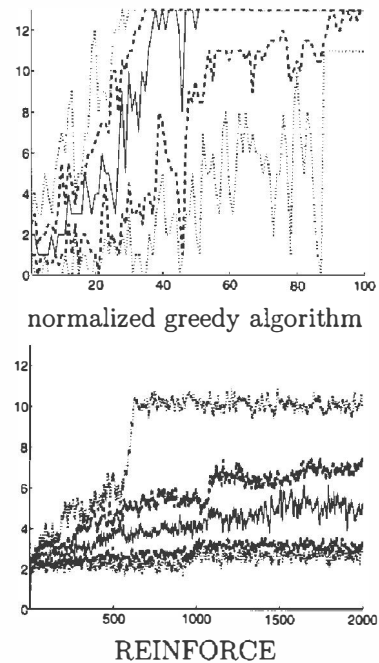

normalized greedy algorithm

REINFORCE

Figure 6: Comparison of the greedy algorithm with a normalized estimator to standard REINFORCE on the load-unload problem. Plotted are the returns as a function of trial number for ten runs of the algorithm. In the case of REINFORCE, the returns have been smoothed over ten trials. The center line is the median of the returns. The lines on either side are the first and third quartiles. The top and bottom lines are the minimum and maximum values.

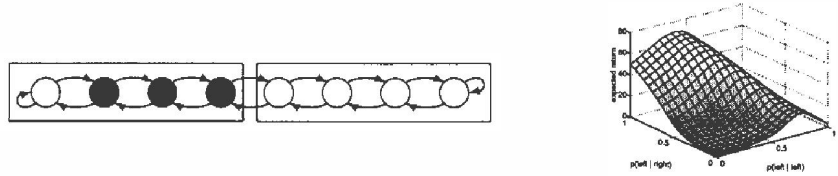

Figure 3: Left: Diagram of the "left-right" world. This world has eight states. The agent receives no reward in the outlined states and one unit of reward each time it enters one of the solid states. The agent only observes whether it is in the left or right set of boxed states (a single bit of information). Each trial begins in the fourth state from the left and lasts 100 time steps. Right: The true expected return as a function of policy for this world.

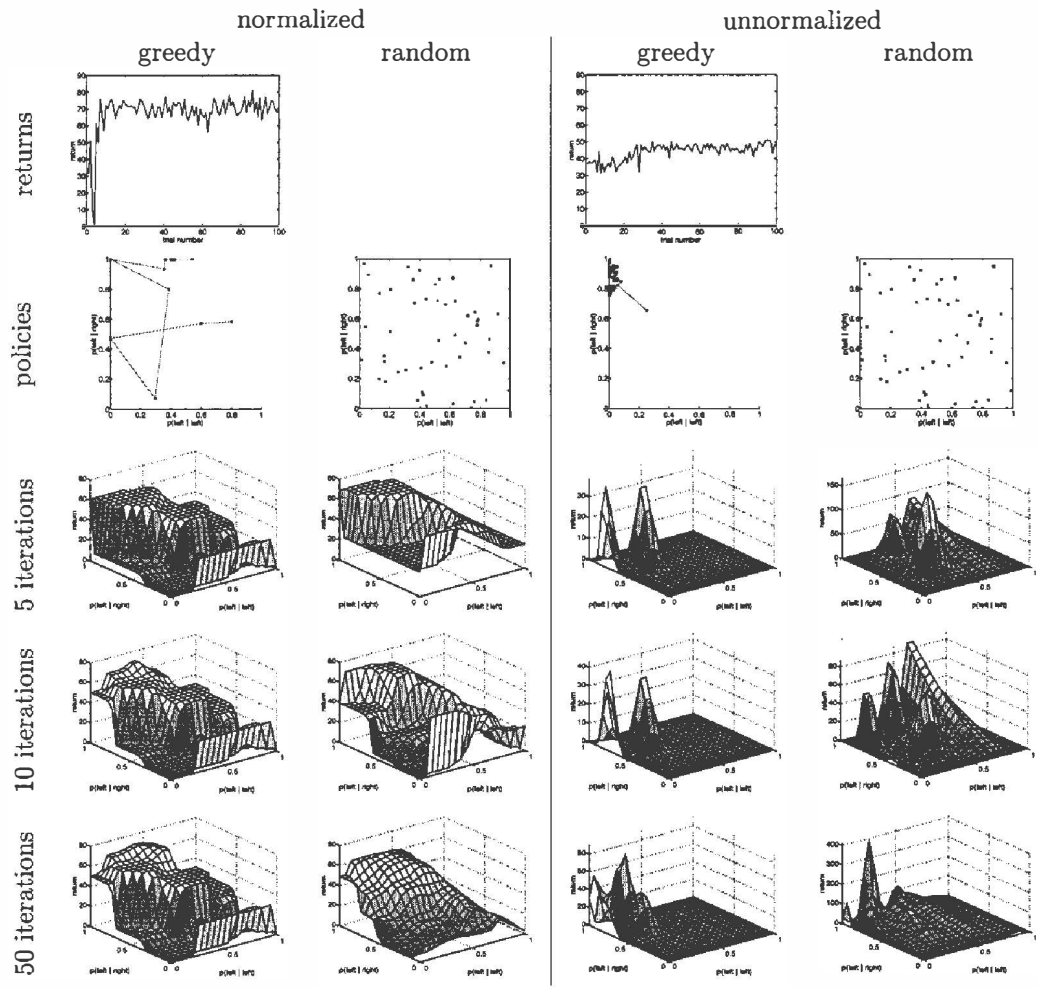

Figure 4: A comparison of the normalized and unnormalized estimators for single set of observations. For each estimator, the return estimates are shown plotted after 5, 10, and 50 iterations (samples). The left column is for the greedy policy improvement algorithm and the right column is for uniformly sampled policies. The first row shows the returns as a function of trial number. The second shows the path taken in policy space (or, for right columns, the random samples taken). Both estimators were given the same sequence of data for the random case. The random sampling of policies produces a better return surface in general, whereas the greedy algorithm quickly maximizes the return (within 10 episodes) and provides a better estimate of the surface near the maximum.

FORCE in figure 6. The REINFORCE algorithm frequently gets stuck in local minima. The graph shown is for the best settings for the step size schedule and bias term of REINFORCE. In the best case, REINFORCE converges to a near optimal policy in around 500 trials The greedy algorithm run with the normalized estimate makes much better use of the data. Not only does it reuse old experience, it has an explicit model of the memory bit and therefore does not need to learn the "dynamics" of the memory. Most runs converge to the optimal policy in about 50 trials. One trial took about twice as long to converge to a slightly suboptimal policy. Although not shown here, if we remove REINFORCE's memory model disadvantage by requiring both algorithms to use "external" memory, the importance sampling algorithm's performance degrades by approximately a factor of 2.

## 5   Conclusion

We think this normalized estimator shows promise. It makes good use of the data and when combined with a greedy algorithm produces quick learning. We would like to extend it in two immediate ways. The first is to provide error estimates or bounds on the return estimate. Although we have a formula for the variance of the estimator, we still need a good estimate of this variance from the samples (the formula requires full knowledge of the POMDP). Such an estimate would allow for exploration to be incorporated into the algorithm. Second, the estimate needs to be "sparsified." After $n$ trials, computing the estimate (or its derivative) for a given policy takes $O(n)$ work. This makes the entire algorithm quadratic in the number of trails. However, a similar estimate could probably be achieved with fewer points. Remembering only the important trials would produce a simpler estimate.

Finally, it may seem disturbing that we must remember which policies were used on each trail. The return doesn't really depend on the policy that the agent wants to execute; it only depends on how the agent actually does act. In theory we should be able to forget which policies were tried; doing so would allow us to use data which was not gathered with a specified policy. The policies are necessary in this paper as proxies for the unobserved state sequences. We hope in future work to remove this dependence.

### Acknowledgements

This report describes research done within CBCL in the Department of Brain and Cognitive Sciences and in the AI Lab at MIT. This research is sponsored by a grants from ONR contracts Nos. N00014-93-1-3085 & N00014-95-1-0600, and NSF contracts Nos. IIS-9800032 & DMS-9872936. Additional support was provided by: AT&T, Central Research Institute of Electric Power Industry, Eastman Kodak Company, Daimler-Chrysler, Digital Equipment Corporation, Honda R&D Co., Ltd., NEC Fund, Nippon Telegraph & Telephone, and Siemens Corporate Research, Inc.

## References

[Geweke, 1989] Geweke, J. (1989). Bayesian inference in econometric models using monte carlo integration. *Econometrica*, 57(6):1317–1339.

[Hesterberg, 1995] Hesterberg, T. (1995). Weighted average importance sampling and defensive mixture distributions. *Technometrics*, 37(2):185–194.

[Kearns et al., 1999] Kearns, M., Mansour, Y., and Ng, A. (1999). Approximate planning in large POMDPs via reusable trajectories. In *Advances in Neural Information Processing Systems*, pages 1001–1007.

[Kloek and van Dijk, 1978] Kloek, T. and van Dijk, H. K. (1978). Bayesian estimates of equation system parameters: An application of integration by monte carlo. *Econometrica*, 46(1):1–19.

[Meuleau et al., 2001] Meuleau, N., Peshkin, L., and Kim, K.-E. (2001). Exploration in gradient-based reinforcement learning. Technical Report AI-MEMO 2001-003, MIT, AI Lab.

[Peshkin et al., 1999] Peshkin, L., Meuleau, N., and Kaelbling, L. P. (1999). Learning policies with external memory. In *Proceedings of the Sixteenth International Conference on Machine Learning*.

[Peshkin and Mukherjee, 2001] Peshkin, L. and Mukherjee, S. (2001). Bounds on sample size for policy evaluation in markov environments. In *Fourteenth Annual Conference on Computational Learning Theory*.

[Precup et al., 2001] Precup, D., Sutton, R. S., and Dasgupta, S. (2001). Off-policy temporal-difference learning with function approximation. In *Proceedings of the Eighteenth International Conference on Machine Learning*.

[Precup et al., 2000] Precup, D., Sutton, R. S., and Singh, S. (2000). Eligibility traces for off-polcy policy evaluation. In *Proceedings of the Seventeenth International Conference on Machine Learning*.

[Press et al., 1992] Press, W. H., Teukolsky, S. A., Vetterling, W. T., and Flannery, B. P. (1992). *Numerical Recipes in C*. Cambridge University Press, second edition.

[Rubinstein, 1981] Rubinstein, R. Y. (1981). *Simulation and the Monte Carlo Method*. John Wiley & Sons.

[Shelton, 2001] Shelton, C. R. (2001). Policy improvement for POMDPs using normalized importance sampling. Technical Report AI-MEMO 2001-002, MIT, AI Lab.

[Williams, 1992] Williams, R. J. (1992). Simple statistical gradient-following algorithms for connectionist reinforcement learning. *Machine Learning*, 8:229–256.